\documentclass{article}

\usepackage{microtype}
\usepackage{graphicx}
\usepackage{subcaption}
\usepackage{booktabs} 
\usepackage[table]{xcolor}

\usepackage{hyperref}

\usepackage[preprint]{icml2026}

\usepackage{amsmath}
\usepackage{amssymb}
\usepackage{mathtools}
\usepackage{amsthm}

\usepackage[capitalize,noabbrev]{cleveref}

\theoremstyle{plain}

\theoremstyle{definition}

\theoremstyle{remark}

\usepackage[textsize=tiny]{todonotes}

\icmltitlerunning{Surgical Attention Tracking}

\begin{document}

\twocolumn[
  \icmltitle{SurgAtt-Tracker: Online Surgical Attention Tracking via Temporal Proposal Reranking and Motion-Aware Refinement}



  \icmlsetsymbol{equal}{*}

  \begin{icmlauthorlist}
    \icmlauthor{Rulin Zhou}{equal,cuhk,hku}
    \icmlauthor{Guankun Wang}{equal,cuhk}
    \icmlauthor{An Wang}{equal,cuhk}
    \icmlauthor{Yujie Ma}{szu}
    \icmlauthor{Lixin Ouyang}{szu}
    \icmlauthor{Bolin Cui}{cuhk}
    \icmlauthor{Junyan Li}{szu}
    \icmlauthor{Chaowei Zhu}{szph}
    \icmlauthor{Mingyang Li}{hku}
    \icmlauthor{Ming Chen}{szu}
    \icmlauthor{Xiaopin Zhong}{szu}
    \icmlauthor{Peng Lu}{hku}
    \icmlauthor{Jiankun Wang}{nkd}
    \icmlauthor{Xianming Liu}{szph}
    \icmlauthor{Hongliang Ren}{cuhk}
  \end{icmlauthorlist}

  \icmlaffiliation{cuhk}{Department of Electronic Engineering, The Chinese University of Hong Kong, Hong Kong SAR, China}
  \icmlaffiliation{szph}{Division of Gastrointestinal Surgery, Shenzhen People's Hospital, China}
  \icmlaffiliation{szu}{College of Mechatronics and Control Engineering, Shenzhen University, China}
  \icmlaffiliation{hku}{Department of Mechanical Engineering, The University of Hong Kong, Hong Kong SAR, China}
  \icmlaffiliation{nkd}{Department of Electronic and Electrical Engineering, Southern University of Science and Technology}

  \icmlcorrespondingauthor{Hongliang Ren}{hlren@ee.cuhk.edu.hk}


  \vskip 0.3in
]



\printAffiliationsAndNotice{} 

\begin{abstract}
Accurate and stable field-of-view (FoV) guidance is critical for safe and efficient minimally invasive surgery, yet existing approaches often conflate visual attention estimation with downstream camera control or rely on direct object-centric assumptions. In this work, we formulate surgical attention tracking as a spatio-temporal learning problem and model surgeon focus as a dense attention heatmap, enabling continuous and interpretable frame-wise FoV guidance. We propose SurgAtt-Tracker, a holistic framework that robustly tracks surgical attention by exploiting temporal coherence through proposal-level reranking and motion-aware refinement, rather than direct regression. To support systematic training and evaluation, we introduce SurgAtt-1.16M, a large-scale benchmark with a clinically grounded annotation protocol that enables comprehensive heatmap-based attention analysis across procedures and institutions. Extensive experiments on multiple surgical datasets demonstrate that SurgAtt-Tracker consistently achieves state-of-the-art performance and strong robustness under occlusion, multi-instrument interference, and cross-domain settings. Beyond attention tracking, our approach provides a frame-wise FoV guidance signal that can directly support downstream robotic FoV planning and automatic camera control. 
\end{abstract}

\section{Introduction}

\begin{figure}[ht]
\centering
\centerline{\includegraphics[width=\columnwidth]{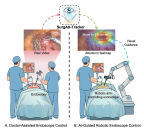}}
    \caption{
    SurgAtt-Tracker enables AI-guided endoscope control by predicting an attention heatmap from raw endoscopic video.    }
    \label{fig:manuscript}
\end{figure}

Minimally Invasive Surgery (MIS) has revolutionized modern clinical practice across diverse specialties, ranging from general surgery to gynecology and urology~\citep{dupont2021decade}. By performing sophisticated procedures through small incisions guided by endoscopic imaging, laparoscopic surgery offers profound clinical benefits, including minimized intraoperative blood loss, reduced post-operative pain, and accelerated patient recovery compared to traditional open surgery~\citep{taylor2016medical, dupont2022continuum}. Centrally, the laparoscope provides the primary perceptual interface between the surgeon and the operative field, making Field-of-View (FoV) quality critical to surgical safety and precision. However, precise visual guidance remains challenging, as traditional manual control relies on coordinated interaction between the surgeon and a dedicated assistant~\citep{merola2002comparison}. Prolonged procedures often lead to physical fatigue for the assistant, which manifests as image instability, unintended drifting of the FoV, or loss of target tracking. Furthermore, communication delays between the surgeon and assistant often lead to misplaced views that do not meet the surgeon's immediate needs~\citep{fang2024force, gao2025human}. These issues disrupt the surgical flow and increase operative risks, ultimately undermining the benefits of the minimally invasive approach.

To address the limitations of manual control, extensive studies have investigated automated or semi-automated FoV control strategies. Approaches based on explicit external inputs utilize cues such as gaze~\citep{fujii2018gaze} and voice commands~\citep{sandoval2021towards} to directly control camera motion. While intuitive, these interaction-heavy paradigms inevitably increase the surgeon’s cognitive load and may interrupt the surgical workflow. In contrast, some researchers explore internal surgical cues, most notably instrument motion, as implicit indicators of surgeon intent, enabling more natural FoV adjustment without additional user interaction~\citep{gruijthuijsen2022robotic}. Prior approaches proposed a cognitive robotic endoscope that integrates a learned camera quality classifier with predefined policies to select optimal viewpoints~\citep{bihlmaier2014automated}. Subsequent strategies include centering the FoV on a single instrument~\citep{yang2019adaptive}, enclosing multiple tools within a predefined region~\citep{huang2022surgeon}, or adjusting viewpoints via model-based visual servoing~\citep{zhang2023visual}. To improve adaptability across surgical phases, probabilistic models have been introduced to encode surgeon preferences from historical data~\citep{li2021data}. Notably, Li et al.~\citep{li2024gmm} proposed a heuristic decision framework that extracts domain knowledge from clinical videos and formulates FoV control as a constrained optimization problem. Despite their robustness and explainability, these methods typically assume that surgeon attention can be approximated by instrument distributions, which may fail in scenarios involving multiple objects and rapid focus shifts. 

More recently, data-driven and learning-based approaches have been explored to enhance the intelligence of FoV control. These include imitation learning~\citep{rivas2019transferring, li20223d} and attention-aware models that associate surgical actions with camera motion~\citep{gao2022savanet}. Parallel to this trend, human–AI collaborative frameworks have been proposed to balance autonomy and surgeon control, combining lightweight commands with scene understanding to infer intent-aligned viewpoints~\citep{gao2025human}. Although previous works have explored various cues for guiding automated FoV adjustment, none explicitly formulated the surgeon’s visual attention as a dense, spatially distributed heatmap (as shown in Fig.~\ref{fig:manuscript}), which can be used as a continuous and spatially expressive signal for FoV guidance beyond discrete directional cues.

Existing video-level heatmap visualizations~\citep{droste2020unified, xu2025etsm} are usually used for post-hoc analysis or monitoring rather than real-time attention modeling rooted in the surgeon's latent perceptual priorities. Rather than directly predicting camera motion or prescribing FoV control policies, we argue that a prerequisite for intelligent endoscopic assistance is the ability to reliably track where the surgeon is visually attending over time. In practice, surgical attention is not a static object nor a single instrument, but a context-dependent perceptual target that shifts across tissues, tools, and procedural stages~\citep{wang2026automatic}. It may focus on an active instrument during dissection, transition to exposed tissue during inspection, or rapidly relocate in response to bleeding or unexpected events. This observation motivates us to decouple attention modeling from downstream camera control and to treat surgical attention tracking as a learning problem. We represent attention as a dense spatial heatmap that captures graded focus intensity, rather than a single point or bounding box, enabling nuanced modeling of distributed and smoothly evolving visual priorities. However, learning to track surgical attention in real time presents several unique challenges. First, attention is a latent cognitive state that lacks explicit visual markers, making direct supervision infeasible. Second, frame-wise detectors often rank attention-relevant regions inconsistently due to occlusion, motion blur, and low inter-class separability, leading to unstable Top-1 predictions. Third, attention exhibits strong temporal coherence yet can undergo abrupt shifts, requiring models to balance short-term consistency with responsiveness. 

To address these challenges, we introduce SurgAtt-Tracker, a framework that explicitly tracks surgical attention by exploiting temporal coherence among ambiguous detector proposals. SurgAtt-Tracker formulates attention tracking as a proposal reranking and refinement problem conditioned on cross-frame consistency, enabling stable dense attention heatmap prediction without relying on direct regression. To support this formulation, we propose SurgAtt-1.16M, a large-scale benchmark for surgical attention tracking spanning diverse laparoscopic procedures. Built upon a clinically grounded annotation protocol that bridges discrete expert attention regions and continuous heatmap representations, SurgAtt-1.16M enables systematic learning and evaluation of temporally coherent surgical attention. 

In summary, our contributions are fourfold: (i) We formulate surgical attention tracking as a spatio-temporal learning problem and propose dense heatmap as its core modeling primitive; (ii) We propose SurgAtt-Tracker, a holistic framework that robustly tracks attention by exploiting proposal-level reranking and motion-aware refinement; (iii) We introduce SurgAtt-1.16M, a large-scale benchmark that enables systematic heatmap attention evaluation and generalization analysis across procedures and institutions. (iv) Extensive experiments across multiple datasets demonstrate state-of-the-art performance and strong robustness of our  SurgAtt-Tracker under occlusion, multi-instrument interference, and cross-domain settings.

\begin{figure}[t]
  \begin{center}
\centerline{\includegraphics[width=\linewidth]{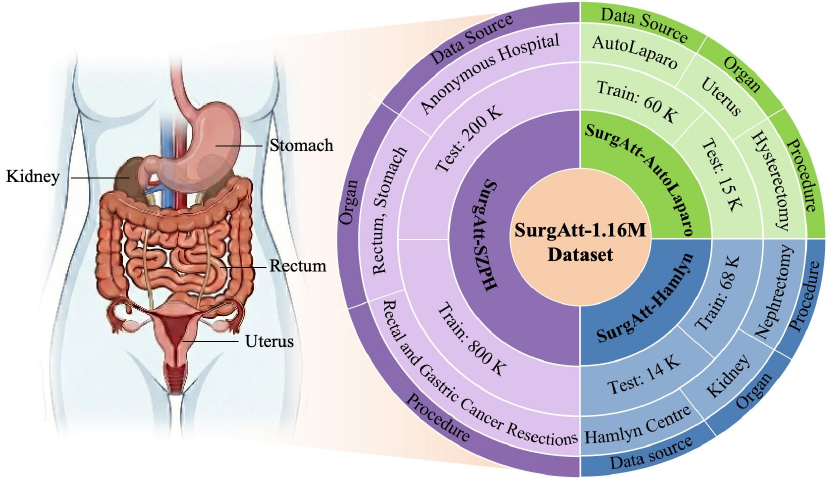}}
    \caption{
      Overview of the SurgAtt-1.16M dataset, illustrating its anatomical coverage, data sources, and unified organization across organs, procedures, and annotation types.
    }
    \label{fig:overall_dataset}
  \end{center}
   \vskip -0.5in
\end{figure}


\section{Surgical Attention Tracking: Task and Dataset}

To advance surgical FoV control from heuristic instrument tracking to intent alignment, we formalize the task of Surgical Attention Tracking (SAT) and introduce \textbf{SurgAtt-1.16M}, a large-scale benchmark for systematic training and evaluating SAT under a unified protocol.

\subsection{Task Formulation}

We define SAT as a spatio-temporal density estimation problem. Let $\mathcal{V} = \{I_1, I_2, ..., I_T\}$ be a surgical video sequence. At each frame $I_t$, the primary goal is to predict the primary focus of the surgeon's operation. Unlike standard object tracking, surgical attention is not a rigid object but a latent cognitive state inferred from visual cues. 
To bridge the gap between discrete surgical interactions and continuous visual focus, we represent attention using two complementary modalities: 
First, we define the \textbf{Latent Attention Center ($B_t$)} as a bounding box localizing the primary interaction (e.g., tissue-tool contact), serving as a discrete anchor for training stability and camera centering. Second, we model \textbf{Attention Density ($\mathbf{H}_t$)} as a dense heatmap to capture the spatial spread and uncertainty of focus; this continuous representation offers a more nuanced tracking target than binary boxes and serves as the primary output for evaluation.

\subsection{Data Construction: The SurgAtt-SZPH Subset}

Existing public datasets often lack the complexity required to model expert attention shifts. A core contribution of this work is the curation of \textbf{SurgAtt-SZPH}, a high-quality clinical dataset derived from 141 hours of laparoscopic footage across 25 patients (Rectum/Stomach resections). 
Crucially, valid attention modeling requires high operational density. Randomly sampled surgical frames are often static. We therefore employed a two-stage filtering pipeline: (1) \textit{Motion filtering} using RAFT to discard static intervals ($<$20 hours retained), followed by (2) \textit{Expert clinical review} where surgeons screened segments based on four criteria: (i) operational complexity, (ii) coverage of common surgical instruments, (iii) inclusion of representative procedures, and (iv) correctness of surgical operations. This yielded 125 interaction-rich clips (approx. 1.12M frames) lasting 4--8 minutes each. Details are in \textbf{Appendix~\ref{app:dataset_appendix}}.

\subsection{Annotation Protocol: From Discrete Intent to Continuous Attention}
\label{sec:annotation_protocol}
Defining ground truth (GT) for latent attention remains challenging: eye-tracking signals are often noisy and reactive, while raw instrument detections fail to capture surgical intent. Thus, we propose a hierarchical annotation strategy that derives continuous attention from discrete expert rules.

\textbf{Step 1: Hierarchical Intent Labeling $(B_t)$.} 
Three surgical assistants annotated the primary target box $B_t$ for every frame following a strict clinical hierarchy: 
(i) \textit{Tissue-Tool Interaction:} The geometric center of the contact area where the instrument actively engages with tissue (e.g., dissection, suturing) takes highest priority; 
(ii) \textit{Active Effector Dominance:} In multi-instrument scenes, the tool performing the primary maneuver (e.g., electrocautery) takes precedence over passive retractors or static grippers; 
(iii) \textit{Navigational Centering:} During camera movement or global inspection (no active tool action), the attention target defaults to the anatomical center of the FoV.

\textbf{Step 2: Modeling Visual Persistence $(\mathbf{H}_t)$.} 
Human attention exhibits temporal persistence rather than instantaneous disappearance. To model this, we avoid directly rasterizing $B_t$ and instead model attention as a \textit{temporally decaying Gaussian process}. Each discrete box $B_t$ is mapped to an anisotropic Gaussian kernel and accumulate them over time using exponential decay. This transforms the jittery discrete sequence into a smooth, probability-like density map $\mathbf{H}_t$. The mathematical derivation and parameters for this conversion are detailed in \textbf{Appendix~\ref{app:heatmap_generation}}.

\subsection{Benchmark Composition}
We integrate SurgAtt-SZPH with two public datasets, \textbf{AutoLaparo} (Gynecology)~\citep{wang2022autolaparo} and \textbf{Hamlyn} (Urology)~\citep{hamlyn_centre_vision}, adapting their labels to our unified heatmap protocol to form the full \textbf{SurgAtt-1.16M} benchmark. As detailed in Fig.~\ref{fig:overall_dataset}, the benchmark comprises \textbf{1.16 million frames}, offering the largest and most diverse testbed for surgical attention tracking to date, spanning multiple organs and procedures.

\section{Method}

\subsection{Overview}

We conceptualize Surgical Attention Tracking as a \textit{temporally conditioned proposal evolution} problem instead of isolated frame-wise detection. Standard detectors often fail to maintain tracking stability because surgical attention lacks rigid visual features, leading to jittery confidence scores and frequent ID switches. However, a key observation is that while the Top-1 detection prediction is unstable, the true attention target is consistently present within the Top-$K$ proposal set (high recall). To exploit this, \textbf{SurgAtt-Tracker} decouples localization into three cohesive stages: hypothesis generation, temporal reranking, and geometric refinement.

\begin{figure*}[ht]
  \begin{center}
\centerline{\includegraphics[width=0.9\textwidth]{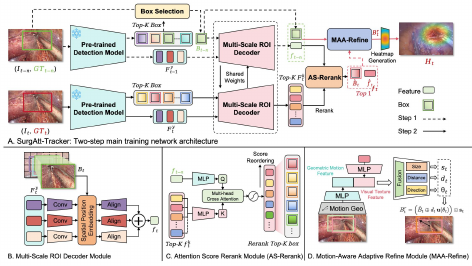}}
    \caption{
      Overview of SurgAtt-Tracker. A frozen detector produces Top-$K$ proposals and multi-scale pyramid features, which are converted into box-aligned embeddings by the Multi-Scale ROI Decoder (B); the AS-Rerank module performs temporal proposal reranking to select the Top-1 attention region (C), and MAA-Refine further refines it using motion-aware geometry (D) and visual evidence to yield the attention heatmap $H_t$.
    }
    \label{fig:main_method}
  \end{center}
  \vskip -0.3in
\end{figure*}

As illustrated in~\cref{fig:main_method}-A, the framework operates on a target frame $I_t$ conditioned on a reference frame $I_{t-n}$ that represents the previous attention state. Initially, a pre-trained, frozen detector acts as a generic region proposer (Sec.~\ref{sec:frozen_detector}), projecting the image into a high-recall discrete search space (Top-$K$ boxes) and and extract multi-scale features with the  \textbf{Multi-Scale ROI Decoder}. 

Subsequently, the \textbf{Attention Score Rerank} module (Sec.~\ref{sec:as_rerank}) discards noisy detection confidence in favor of temporal consistency through the cross-attention mechanism. 
Finally, to overcome the discretization error of fixed proposals, the \textbf{Motion-Aware Adaptive Refine} module (Sec.~\ref{sec:maa_refine}) fuses visual cues with geometric motion history to predict continuous corrections, yielding the precise final box $B_t^r$.

\subsection{Architecture Details}

\subsubsection{Frozen Detector \& Proposal Generation}
\label{sec:frozen_detector}

Our framework begins by transforming the global localization problem into a tractable discrete selection task. We employ a detector $D(\cdot)$ pre-trained on SurgAtt-SZPH as a \textit{high-recall hypothesis generator}. 
Crucially, we maintain $D(\cdot)$ in a \textbf{frozen state} throughout tracking training. This decouples feature extraction from tracking dynamics, preventing the model from overfitting to specific detector failure modes while ensuring stable feature distributions for cross-frame matching.

Given an input frame $I_t$, this module outputs two distinct primitives. First, it defines a \textbf{Discrete Search Space ($\mathcal{B}_t$)}, extracting the Top-$K$ proposals $\mathcal{B}_t=\{B_t^k\}_{k=1}^{K}$ based on raw detection confidence. While the Top-1 box is often unstable due to visual ambiguity, empirical analysis confirms that the true attention target is consistently contained within this expanded candidate set (high recall). Second, to handle the varying scale of surgical interactions, we extract \textbf{Multi-Scale Semantics ($\mathcal{F}_t^{y}$)} in the form of feature pyramids $\mathcal{F}_t^{y}=\{F_t^{(s)}\}_{s\in\{3,4,5\}}$ from the detector neck. These embeddings provide the robust visual descriptors required for the subsequent attention-based retrieval.


\subsubsection{Multi-Scale ROI Decoder}
\label{sec:msr}

Effective temporal reranking demands a robust proposal-level representation that is discriminative and spatially aligned. In surgical scenes, the scale of attention targets varies drastically, ranging from minute instrument tips during dissection to large anatomical regions during inspection, which renders single-scale feature extraction insufficient. To address this, the \textbf{Multi-Scale ROI Decoder (MSR)} acts as a feature transducer, mapping the discrete bounding box candidates back into the continuous semantic space of the frozen detector's feature pyramid $\mathcal{F}_t^{y}$.

As shown in~\cref{fig:main_method}-B, for each proposal $B$, we extract semantics across spatial resolutions to simultaneously capture fine-grained texture and global context. Specifically, we employ ROIAlign to harvest fixed-size features from each pyramid level proportional to the box coordinates. These multi-resolution descriptors are projected via a scale-specific layer $\phi_s(\cdot)$ with positional encoding and fused via summation to yield a unified, box-aligned embedding $f_t(B)$:
\begin{equation}
f_t(B)=\sum_{s\in\{3,4,5\}} \mathrm{Align}\!\left(\phi_s\!\left(F_t^{(s)}\right),\, \pi_s(B)\right),
\label{eq:msr}
\end{equation}
where $\pi_s(\cdot)$ maps $B$ from image coordinates to the coordinate system of the $s$-th feature map, and $\mathrm{Align}(\cdot)$ denotes the interpolation-based ROI feature extraction operator. This process ensures that subsequent tracking modules operate on consistent, scale-invariant visual descriptors regardless of the target's changing size.

\subsubsection{Attention Score Rerank Module}
\label{sec:as_rerank}

While the frozen detector provides discriminative feature pyramids, its Top-$K$ proposal ranking relies solely on static, per-frame confidence scores, which are often unstable under surgical conditions such as occlusion, motion blur, and rapid viewpoint changes. To address this, we introduce the \textbf{Attention Score Rerank (AS-Rerank)} module, which shifts the selection criterion from static detection confidence to dynamic temporal consistency. Given that surgical attention exhibits pronounced trajectory coherence, this module identifies the optimal candidate at time $t$ by measuring its compatibility with a trusted reference state.

As illustrated in~\cref{fig:main_method}-C, AS-Rerank operates as a cross-frame retrieval mechanism. It takes the trusted reference ROI embedding $f_r$ (from frame $I_r$) and the set of current proposal embeddings $\{f_t^k\}_{k=1}^{K}$ (from frame $I_t$) as input. We employ a multi-head cross-attention mechanism to compute a scalar reranking logit $s_t^k$ for each proposal:
\begin{equation}
s_t^k = \mathrm{Att}\!\left(f_r,\, f_t^k\right),
\label{eq:as_rerank_logit}
\end{equation}
where $\mathrm{Att}(\cdot)$ is a learnable matching function that weighs feature similarity against the historical context. Finally, we select the index $\hat{k}_t=\arg\max_k s_t^k$ that maximizes this temporal affinity, outputting the Top-1 hypothesis proposal $\hat{B}_t=B_t^{\hat{k}_t}$ and its corresponding feature $\hat{f}_t$ for subsequent geometric refinement.

\subsubsection{Motion-Aware Adaptive Refine Module}
\label{sec:maa_refine}

Although the AS-Rerank module successfully identifies the most plausible candidate from the proposal pool, the selected box $\hat{B}_t$ remains quantized by the detector's anchor grid, often leading to residual misalignment in center positioning and scale. To transcend this discretization limit, the \textbf{Motion-Aware Adaptive Refine (MAA-Refine)} module performs continuous geometric correction (\cref{fig:main_method}-D). Crucially, visual-only refinement is often ambiguous in low-texture surgical scenes, so we condition the correction on explicit {motion prior} encoded from the reference state.

The module fuses two inputs: the visual semantic embedding $\hat{f}_t$ of the selected Top-1 proposal, and a geometric motion descriptor $g_t=\mathrm{Geo}(\hat{B}_t, B_r^{s})$. Here, $B_r^{s}$ represents the trusted attention state from the reference frame $I_{t-n}$ (using the previous prediction $\hat{B}_{t-1}$ during inference). The descriptor $g_t$ explicitly encodes the spatial evolution of the target, capturing normalized coordinates, relative center displacement, and log-scale aspect variations. By concatenating $\hat{f}_t$ and $g_t$ into a lightweight MLP, the network learns to adjust the box based on both local tissue texture and the momentum of the surgeon's attention.

Unlike standard offset regression, which predicts unconstrained Cartesian shifts, we adopt a polar-based formulation that decouples directionality from magnitude. The network predicts a correction tuple consisting of a displacement direction $\theta_t$, a scalar magnitude $d_t$, and channel-wise scale factors $\mathbf{s}_t=(s_t^{w},s_t^{h})$. Let $\psi(\hat{B}_t)=(c_x, c_y, w, h)$ denote the parameterization of the coarse proposal. We update the center via a directed step $d_t$ along the unit vector $\mathbf{u}(\theta_t)=(\cos\theta_t,\sin\theta_t)$, and modulate the dimensions via multiplicative scaling:
\begin{equation}
\begin{aligned}
(c_x^{r},c_y^{r}) &= (c_x,c_y) + d_t\,\mathbf{u}(\theta_t),\\
(w^{r},h^{r}) &= (w,h)\odot (s_t^{w},s_t^{h}),
\end{aligned}
\label{eq:maa_update}
\end{equation}
where $\odot$ denotes element-wise multiplication. The final refined box $B_t^{r}$ is reconstructed from these parameters and serves as the precise anchor for generating the continuous attention heatmap $\mathbf{H}_t$.


\subsection{Training Objectives}
\label{sec:loss}

To jointly optimize discrete proposal selection and continuous geometric refinement, our training objective is composed of two hierarchical components: a \textbf{reranking loss} that enforces correct ordering within the hypothesis pool, and a \textbf{refinement loss} that regularizes the precise regressors. The total objective is defined as:
\begin{equation}
\mathcal{L}_{\text{total}}
=
\mathcal{L}_{\text{rerank}}^{(t)}
+\mathcal{L}_{\text{refine}}^{(t)}.
\label{eq:loss_total}
\end{equation}
We compute these losses on the target frame $I_t$, isolating valid proposals (those with non-zero overlap) to prevent noise from background outliers.

\subsubsection{Reranking losses}

The goal of reranking is to assign high scores to proposals that are spatially closest to the GT $B_t^{\mathrm{gt}}$. Let $\mathcal{B}_t=\{B_t^k\}_{k=1}^{K}$ be the set of valid proposals and $s_t=\{s_t^k\}_{k=1}^{K}$ be the predicted logits. We identify the optimal proposal index $k_t^\star$ as the one minimizing the center distance to the GT:
\begin{equation}
k_t^\star
=
\arg\min_{k} \left\|c(B_t^k)-c(B_t^{\mathrm{gt}})\right\|_2.
\end{equation}
To ensure robust ranking, we employ a multi-task objective combining hard classification, soft geometric regularization, and listwise ranking:

\textbf{1. Hard Top-1 Classification ($\mathcal{L}_{\text{ce}}$).}
We maximize the probability of selecting the single best candidate $k_t^\star$ using standard cross-entropy:
\begin{equation}
\mathcal{L}_{\text{ce}}
=
\mathrm{CE}\!\left(\mathrm{softmax}(s_t),\, k_t^\star\right).
\end{equation}


\textbf{2. Soft Geometric Regularization ($\mathcal{L}_{\text{geo}}$).}
Hard classification ignores the spatial distribution of the prediction. To inject geometric awareness, we compute a \textit{soft-aggregated box} $\bar{B}_t$ by weighing all proposals using their predicted probabilities $\pi_t^k = \frac{\exp(s_t^k/\tau)}{\sum_j\exp(s_t^j/\tau)}$ (with temperature $\tau{=}0.15$). We then minimize the Huber loss between the center of this aggregated box and the GT:
\begin{equation}
\mathcal{L}_{\text{geo}}
=
\mathcal{L}_{\text{Huber}}\!\left(
\| c(\bar{B}_t) - c(B_t^{\mathrm{gt}}) \|_2
\right),
\end{equation}
which encourages the probability mass to concentrate around the spatial location of the target, smoothing the decision surface.

\textbf{3. Top-$M$ Listwise Ranking ($\mathcal{L}_{\text{rank}}$).}
Standard Top-1 supervision induces a sharp decision boundary that suppresses all non-target proposals equally, regardless of their geometric utility. However, surgical attention is often spatially smooth; proposals slightly offset from the exact center still capture valid context. To exploit this, we enforce a local ranking constraint: the model's confidence distribution should mirror the geometric error distribution within the elite candidate pool. We select the Top-$M$ ($M{=}5$) proposals closest to the GT and construct a ``geometric teacher'' distribution $q_t$ that decays with spatial error $\epsilon_t$:
\begin{equation}
q_t^k
=
\frac{\exp(-\epsilon_t^k/\sigma)}{\sum_{j}\exp(-\epsilon_t^j/\sigma)}, \quad
p_t^k
=
\frac{\exp(s_t^k)}{\sum_{j}\exp(s_t^j)}.
\end{equation}
By minimizing the cross-entropy between the predicted distribution $p_t$ and this geometric prior $q_t$ (with smoothing factor $\sigma=15$px), i.e., $\mathcal{L}_{\text{rank}}^{(t)} = - \sum_{k} q_t^k \log p_t^k$, we ensure that tracking confidence degrades smoothly rather than abruptly. This provides informative gradients for determining relative quality among high-overlap candidates, preventing the tracker from discarding useful near-miss hypotheses.

Overall, the final reranking loss is the weighted sum:
\begin{equation}
\mathcal{L}_{\text{rerank}}^{(t)}
=
\mathcal{L}_{\text{ce}}
+\lambda_{\text{geo}}\mathcal{L}_{\text{geo}}
+\lambda_{\text{rank}}\mathcal{L}_{\text{rank}},
\end{equation}
where we empirically set $\lambda_{\text{geo}}=0.1$ and $\lambda_{\text{rank}}=0.5$ to balance the relative gradient magnitudes across terms.

\subsubsection{Refinement loss}
\label{sec:loss_refine}

Once the optimal proposal is selected, our goal is to fine-tune its geometry to precisely match the GT $B_t^{\mathrm{gt}}$. We supervise the final refined box $B_t^{r}$ by decoupling the regression task into two orthogonal components: \textit{translation alignment} and \textit{scale calibration}.

\textbf{1. Center Alignment Loss ($\mathcal{L}_{\text{dist}}$).}
To ensure precise localization, we penalize the spatial displacement between the predicted center $c(B_t^{r})$ and the GT center $c(B_t^{\mathrm{gt}})$. We employ a normalized Huber loss to be robust against outliers while maintaining sensitivity to small errors:
\begin{equation}
\mathcal{L}_{\text{dist}}^{(t)}
=
\mathcal{L}_{\text{Huber}}\!\left(
\frac{\|c(B_t^{r})-c(B_t^{\mathrm{gt}})\|_2}{\sqrt{W^2+H^2}}
\right),
\label{eq:loss_dist}
\end{equation}
where the diagonal normalization factor $\sqrt{W^2+H^2}$ ensures the loss magnitude is invariant to image resolution.

\textbf{2. Log-Space Scale Loss ($\mathcal{L}_{\text{scale}}$).}
Direct regression of width and height is sensitive to object scale variance; for instance, a 10-pixel error is negligible for large organs yet catastrophic for small instrument tips. To address this, we regress dimensions in logarithmic space, thereby equalizing relative errors across scales. Let $\mathbf{s}(B) = (w, h)$ be the size vector of box $B$. We define:
\begin{equation}
\mathcal{L}_{\text{scale}}^{(t)}
=
\mathcal{L}_{\text{Huber}}\!\left(
\log \mathbf{s}(B_t^{r}) - \log \mathbf{s}(B_t^{\mathrm{gt}})
\right).
\label{eq:loss_scale}
\end{equation}

The total refinement loss is a weighted combination:
\begin{equation}
\mathcal{L}_{\text{refine}}^{(t)}
=
\lambda_{\text{dist}}\mathcal{L}_{\text{dist}}^{(t)}
+\mathcal{L}_{\text{scale}}^{(t)},
\label{eq:loss_ref}
\end{equation}
where we set $\lambda_{\text{dist}}{=}0.1$ to balance the gradients between translation and scaling tasks.

\section{Experiments and Result Analysis}

\begin{table}[t]
  \caption{Quantitative comparison of attention heatmap prediction on SurgAtt-SZPH dataset using NSS, CC, SIM, MSE, and MAE. }
  \label{tab:heatmap_metrics}
  \vskip -0.1in
  \begin{center}
    \begin{small}
      \begin{sc}
        \resizebox{\linewidth}{!}{
        \begin{tabular}{lccccc}
          \toprule
          Model & NSS$\uparrow$ & CC$\uparrow$ & SIM$\uparrow$ & MSE$\downarrow$ & MAE$\downarrow$ \\
          \midrule
           \rowcolor{orange!12}
          \multicolumn{6}{c}{\textit{U-Net--based}}       \\
          Unet~\citep{ronneberger2015u}             & 1.051    & 0.393   & 0.490    & 0.079    & 0.191    \\
          TransUNet~\citep{chen2021transunet}        & 1.947    & 0.697    & 0.651    & 0.033    & 0.101    \\
          UNeXt~\citep{valanarasu2022unext}            & 1.964    & 0.712    & 0.646    & 0.029    & 0.109    \\
          EMCAD~\citep{rahman2024emcad}           & 2.203    & 0.776    & 0.711    & 0.0261    & 0.089    \\

        \midrule
        \rowcolor{blue!15}
          \multicolumn{6}{c}{\textit{Regression-based}} \\
          SASNet~\citep{song2021choose}           & 1.690    & 0.669    & 0.608    & 0.032    & 0.138    \\
          RCMNet~\citep{xu2025etsm}           & 1.763    & 0.684    & 0.618    & 0.036    & 0.115    \\
          SalFoM~\citep{moradi2024salfom}           & 2.030    & 0.702    & 0.647    & 0.028    & 0.094    \\
          \midrule
          \rowcolor{purple!18}
          \multicolumn{6}{c}{\textit{Object Tracking--based}} \\

          AQATrack~\citep{xie2024autoregressive}         & 1.093    & 0.397    & 0.559        & 0.070        & 0.199        \\
          ODTrack~\citep{zheng2024odtrack}          & 1.343    & 0.212    & 0.572    & 0.068    & 0.193    \\
          SPMTrack-B~\citep{cai2025spmtrack}       & 1.720 & 0.571 & 0.643 & 0.060 & 0.149 \\
        LoRAT-B~\citep{lorat}          & 1.837 & 0.597 & 0.669 & 0.053 & 0.131 \\
        LoRATv2-B~\citep{linloratv2}        & 1.853 & 0.649 & 0.676 & 0.047 & 0.126 \\
        MCITrack~\citep{kang2025exploring}         & 1.979 & 0.691 & 0.690 & 0.041 & 0.107 \\
        
        \midrule
        \rowcolor{green!15}
        \multicolumn{6}{c}{\textit{Detection-based}} \\

        YOLOv11~\citep{khanam2024yolov11}          & 2.195 & 0.796 & 0.721 & 0.030 & 0.099 \\
        YOLOv12~\citep{tian2025yolov12}          & 2.285 & 0.793 & 0.723 & 0.028 & 0.090 \\
        RT-DETR~\citep{lv2023detrs}         & 2.359 & 0.820 & 0.737 & 0.025 & 0.087 \\
    RT-DETRv2~\citep{lv2024rtdetrv2improvedbaselinebagoffreebies}        
    & \underline{2.370} & \underline{0.822} & \underline{0.739} & \underline{0.024} & \underline{0.086} \\
        
        \midrule
        \textbf{SurgAtt-Tracker(Ours)}    & \textbf{2.580} & \textbf{0.871} & \textbf{0.829} & \textbf{0.015} & \textbf{0.051} \\

          \bottomrule
        \end{tabular}
        }
      \end{sc}
    \end{small}
  \end{center}
  \vskip -0.3in
\end{table}

\subsection{Implementation Details}

We group baselines into four categories: (i) U-Net-based, (ii) regression-based, (iii) object-tracking-based, and (iv) detection-based methods. To enable fair comparison, we convert all model outputs into a normalized attention heatmap under a unified evaluation protocol; model-specific adaptations are deferred to the supplementary material.
Following prior works~\citep{droste2020unified, xu2025etsm}, we report five metrics: NSS, CC, SIM, MSE and MAE, where the former three assess distributional alignment and the latter two measure pixel errors. All frames are resized to $960{\times}540$. We use AdamW (lr $1{\times}10^{-4}$) and run all experiments on a single NVIDIA A100. Additional implementation details, including metrics, model-specific adaptations, training schedules, are provided in the \textbf{appendix~\ref{app:experiment_detail}}


\begin{table}[t]
  \caption{Zero-shot and fine-tuning performance on the SurgAtt-AutoLaparo and SurgAtt-Hamlyn datasets.}
  \label{tab:zero_finetune}
  \centering
  \begin{small}
    \begin{sc}
      \setlength{\tabcolsep}{6pt} 
      \renewcommand{\arraystretch}{1.08} 
      \resizebox{\linewidth}{!}{%
      \begin{tabular}{lccccc}
        \toprule
        Model & NSS$\uparrow$ & CC$\uparrow$ & SIM$\uparrow$ & MSE$\downarrow$ & MAE$\downarrow$ \\
        \midrule

        \rowcolor{gray!10}
        \multicolumn{6}{c}{\textbf{SurgAtt-AutoLaparo}} \\
        \midrule
        \rowcolor{cyan!12}
        \multicolumn{6}{c}{\textit{Zero-shot}} \\

        EMCAD                     & 1.378 & 0.593 & \underline{0.648} & 0.077 & 0.214 \\
        SalFoM                    & 1.408 & 0.531 & 0.594 & 0.078 & 0.206 \\
        MCITrack                  & 1.440 & 0.578 & 0.503 & 0.074 & 0.196 \\
        YOLOv12                   & 1.551 & 0.658 & 0.520 & 0.059 & \underline{0.146} \\
        RT-DETRv2                 & \underline{1.864} & \underline{0.708} & 0.556 & \underline{0.053} & 0.156 \\
        \textbf{SurgAtt-Tracker (Ours)}
                                 & \textbf{1.949} & \textbf{0.721} & \textbf{0.575} & \textbf{0.035} & \textbf{0.088} \\
         \midrule
        \rowcolor{red!12}
        \multicolumn{6}{c}{\textit{Fine-tuning}} \\
        YOLOv12                   & 2.516 & 0.756 & 0.671 & 0.022 & 0.065 \\
        RT-DETRv2                 & \underline{2.715} & \underline{0.773} & \underline{0.708} & \underline{0.019} & \underline{0.058} \\
        \textbf{SurgAtt-Tracker (Ours)}
                                 & \textbf{2.741} & \textbf{0.837} & \textbf{0.769} & \textbf{0.014} & \textbf{0.048} \\
        \midrule

        \rowcolor{gray!10}
        \multicolumn{6}{c}{\textbf{SurgAtt-Hamlyn}} \\
        \midrule
        \rowcolor{cyan!12}
        \multicolumn{6}{c}{\textit{Zero-shot}} \\
        EMCAD                     & 1.098 & 0.552 & 0.603 & 0.211 & 0.359 \\
        SalFoM                    & 1.133 & 0.497 & 0.574 & 0.164 & 0.235 \\
        MCITrack                  & 1.035 & 0.398 & 0.533 & 0.275 & 0.447 \\
        YOLOv12                   & 1.171 & 0.575 & \underline{0.640} & \underline{0.081} & 0.198 \\
        RT-DETRv2                 & \underline{1.269} & \underline{0.651} & 0.594 & 0.178 & \underline{0.180} \\
        \textbf{SurgAtt-Tracker (Ours)}
                                 & \textbf{1.693} & \textbf{0.686} & \textbf{0.711} & \textbf{0.046} & \textbf{0.128} \\
        \midrule      
        \rowcolor{red!12}
        \multicolumn{6}{c}{\textit{Fine-tuning}} \\
        YOLOv12                   & 1.476 & 0.685 & 0.694 & 0.057 & 0.162 \\
        RT-DETRv2                 & \underline{1.744} & \underline{0.764} & \underline{0.740} & \underline{0.037} & \underline{0.126} \\
        \textbf{SurgAtt-Tracker (Ours)}
                                 & \textbf{2.195} & \textbf{0.890} & \textbf{0.841} & \textbf{0.016} & \textbf{0.071} \\

        \bottomrule
      \end{tabular}%
      }
    \end{sc}
  \end{small}
  \vskip -0.15in
\end{table}

\begin{figure*}[t]
  \begin{center}
\centerline{\includegraphics[width=0.85\textwidth]{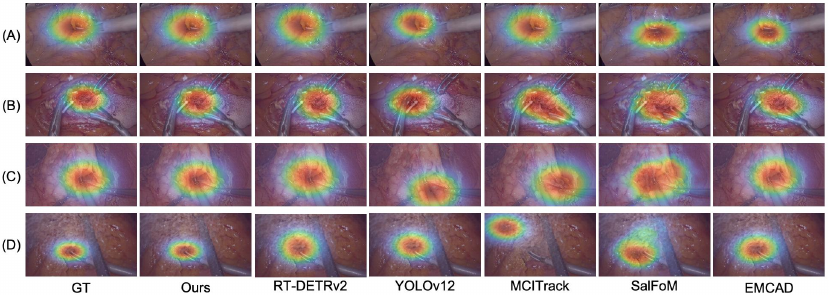}}
    \caption{
      Qualitative comparison of attention heatmaps across diverse surgical scenarios: (A) single-instrument case; (B) multi-instrument without tissue interaction; (C) multi-instrument with tissue interaction; (D) multi-instrument with smoke interference.
    }
    \label{fig:vis_res}
  \end{center}
  \vskip -0.3in
\end{figure*}

\subsection{Main Results}

\paragraph{Evaluation and Visualization on SurgAtt-SZPH.}
We evaluate all baselines and SurgAtt-Tracker on SurgAtt-SZPH, training each model for 4 epochs. SurgAtt-Tracker first pre-trains the YOLOv12 detector for two epochs, then freezes it and optimizes the remaining modules (MSR, AS-Rerank, MAA-Refine) for another 2 epochs under a unified training/inference setting.
\cref{tab:heatmap_metrics} compares attention heatmap prediction on SurgAtt-SZPH across four types of baselines.
U-Net/regression baselines often produce \emph{diffuse} attention with \emph{unstable peaks} and limited temporal coherence, which caps both correlation (NSS/CC/SIM) and pixel-wise errors (MSE/MAE).
This behavior is also evident in~\cref{fig:vis_res}: EMCAD/SalFoM frequently spreads activation over a broad region and shifts the peak away from the GT focus, especially under cluttered tool appearances. Tracking-based methods introduce temporal cues but are primarily optimized for appearance/boundary consistency and can fail under target disappearance or multi-instrument interference (initialized with GT in our evaluation), leading to accumulated drift. As shown in~\cref{fig:vis_res}(B--C), MCITrack is more prone to mis-association when multiple instruments are present, yielding off-target or over-smoothed heatmaps. Detection-based baselines are strongest overall; RT-DETRv2 achieves the best prior performance (NSS=2.370, CC=0.822, SIM=0.739), yet YOLOv12 confidence-driven per-frame ranking remains unreliable under occlusion, blur, and viewpoint changes.
Qualitatively,~\cref{fig:vis_res}(D) further highlights that smoke interference can disrupt per-frame selection, resulting in peak instability or spatial bias even for strong detectors.

In contrast, {SurgAtt-Tracker} consistently outperforms all competitors, achieving (NSS=2.580, CC=0.871, SIM=0.829) with substantially lower errors (MSE=0.015, MAE=0.051).
Compared with RT-DETRv2, our method improves NSS/CC/SIM by 8.9\%/6.0\%/12.2\% and reduces MSE/MAE by 37.5\%/40.7\%. Consistent with these gains,~\cref{fig:vis_res} shows that our predictions remain {sharper} and {better aligned} with GT across diverse scenarios (A--D), including multi-instrument clutter and smoke. Importantly, SurgAtt-Tracker runs at \textbf{12.5 FPS} in an online setting, approaching real-time capability for closed-loop endoscopic control.
We further report per-scene performance on five SurgAtt-SZPH subsets in \textbf{Appendix~\ref{app:szph_scene_radar}}.
Overall, these results validate our formulation of heatmap prediction as temporally consistent retrieval over a high-recall proposal set (AS-Rerank), followed by motion-conditioned continuous refinement (MAA-Refine).

\paragraph{Evaluation on SurgAtt-AutoLaparo and Hamlyn.}
\cref{tab:zero_finetune} evaluates robustness under domain and resolution shifts on two surgical scenarios:
{uterine tumor resection} (AutoLaparo, $1920{\times}1080$) and {nephrectomy} (Hamlyn, $384{\times}192$).
In the \textbf{zero-shot} setting, {SurgAtt-Tracker} achieves strong transfer on both datasets
(AutoLaparo: NSS=1.949, MAE=0.088; Hamlyn: NSS=1.693, MAE =0.128).
After \textbf{fine-tuning}, our performance further improves substantially over our zero-shot model,
with NSS increasing by {+40.6\%}/{+29.7\%} and MAE reduced by {45.5\%}/{44.5\%}
on SurgAtt-AutoLaparo/Hamlyn, respectively, confirming the robustness and transferability of our framework across different pixel videos and scene variations (For more visual results, see the \textbf{appendix~\cref{app:other_data}}).

%

\subsection{Ablation Studies}

\begin{table}[t]
  \caption{Reports detector-only {best-case} performance under three rules (Conf/MinErr/MaxIoU), serving as oracle baselines; SurgAtt-Tracker further incorporates AS-Rerank and MAA-Refine.}
  \label{tab:ablation_rerank_refine}
  \vskip -0.15in
  \begin{center}
    \begin{small}
      \begin{sc}
        \resizebox{\linewidth}{!}{
        \begin{tabular}{lcc|ccccc}
          \toprule
          \textbf{Method}
          & \textbf{Rerank} & \textbf{Refine}
          & \textbf{Err}$\downarrow$ & \textbf{IoU}$\uparrow$
          & \textbf{NSS}$\uparrow$ & \textbf{CC}$\uparrow$ & \textbf{MAE}$\downarrow$ \\
          \midrule
          \rowcolor{orange!10}
          \multicolumn{8}{c}{\textit{Detector-only baselines}} \\
          YOLOv12 (Conf)
          & -- & --
          & 52.192 & 0.589
          & 2.285 & 0.793
          & 0.090 \\
          YOLOv12 (MinErr)
          & -- & --
          & \underline{30.408} & 0.649
          & \underline{2.525} & \underline{0.857}
          & \underline{0.063} \\
          YOLOv12 (MaxIoU)
          & -- & --
          & 32.744 & \underline{0.665}
          & 2.437 & 0.866
          & 0.064 \\
          \midrule
          \rowcolor{green!12}
          \multicolumn{8}{c}{\textit{SurgAtt-Tracker}} \\
          Ours (w/o Refine)
          & \checkmark & \texttimes
          & 36.975 & 0.627
          & 2.313 & 0.823
          & 0.079 \\
          Ours (full)
          & \checkmark & \checkmark
          & \textbf{27.525} & \textbf{0.690}
          & \textbf{2.580} & \textbf{0.871}
          & \textbf{0.051} \\
          \bottomrule
        \end{tabular}
        }
      \end{sc}
    \end{small}
  \end{center}
  \vskip -0.15in
\end{table}

\begin{table}[t]
  \caption{Ablation on the rerank loss weight (Top-$M$).}
  \label{tab:rank_loss_ablation}
  \begin{center}
    \begin{small}
      \begin{sc}
        \resizebox{\linewidth}{!}{
        \begin{tabular}{lccccc}
          \toprule
          Top-$M$ & NSS$\uparrow$ & CC$\uparrow$ & SIM$\uparrow$ & MSE$\downarrow$ & MAE$\downarrow$ \\
          \midrule
          0
            & 1.973
            & 0.729
            & 0.621
            & 0.047
            & 0.144 \\
          3
            & \underline{2.379}
            & \underline{0.859}
            & \underline{0.792}
            & \underline{0.024}
            & \underline{0.086} \\
          5
            & \textbf{2.580}
            & \textbf{0.871}
            & \textbf{0.829}
            & \textbf{0.015}
            & \textbf{0.051} \\
          7
            & 2.195
            & 0.812
            & 0.753
            & 0.033
            & 0.107 \\
          \bottomrule
        \end{tabular}
        }
      \end{sc}
    \end{small}
  \end{center}
  \vskip -0.3in
\end{table}

\begin{table}[t]
  \caption{Impact of temporal gap sampling during training.}
  \label{tab:gap_ablation}
  \begin{center}
    \begin{small}
      \begin{sc}
        \resizebox{\linewidth}{!}{
        \begin{tabular}{lccccc}
          \toprule
          Sample Gap & NSS$\uparrow$ & CC$\uparrow$ & SIM$\uparrow$ & MSE$\downarrow$ & MAE$\downarrow$ \\
          \midrule
          Avg Gap
            & \textbf{2.592}
            & \underline{0.863}
            & 0.788
            & 0.045
            & 0.090 \\
          Random Gap
            & 2.339
            & 0.862
            & \underline{0.798}
            & \underline{0.034}
            & \underline{0.065} \\
          Archer Gap
            & \underline{2.580}
            & \textbf{0.871}
            & \textbf{0.829}
            & \textbf{0.015}
            & \textbf{0.051} \\
          \bottomrule
        \end{tabular}
        }
      \end{sc}
    \end{small}
  \end{center}
  \vskip -0.3in
\end{table}

\paragraph{Ablation on Reranking and Refinement.}
\cref{tab:ablation_rerank_refine} diagnoses \emph{why detector confidence is not sufficient} and verifies the complementary roles of our reranking and refinement.
Although YOLOv12 is capable of producing high-quality candidates, its confidence-based Top-1 selection is often suboptimal:
In contrast, an \emph{oracle} choice within YOLOv12 proposals (MinErr / MaxIoU) reveals that better boxes already exist in the candidate pool (Err = 30.41; IoU = 0.665).

Building on this candidate pool, our AS-Rerank module effectively recovers better hypotheses by reordering proposals according to temporal consistency, reducing localization error from 52.192 to 36.975 and improving NSS from 2.285 to 2.313 (see \textbf{Appendix~\ref{app:ablation}} for detailed analysis of AS-Rerank).
Adding MAA-Refine module reduces Err from 36.98 to 27.53, while consistently boosting heatmap metrics.

\paragraph{Ablation on Top-$M$ Ranking.} We ablate the Top-$M$ setting in the $\mathcal{L}_{\text{rank}}$.
As shown in \cref{tab:rank_loss_ablation}, performance improves from $M{=}0$ to $M{=}5$ but degrades when $M$ is further increased.
Top-5 supervision strikes a good balance between learning meaningful relative orderings and suppressing noisy candidates, yielding the best overall results.

\paragraph{Ablation on Temporal Sample Gap}
Training solely on adjacent pairs ($n{=}1$) may lead to an identity shortcut, encouraging the model to copy the previous prior under slow motion and reducing robustness to rapid tool/camera motion. 
We therefore sample multi-scale temporal gaps during training: for each target frame $t$, we draw $n\!\sim\!\mathrm{Cat}(\mathcal{N},\mathcal{P})$ with $\mathcal{N}=\{1,2,4,8,16,32\}$ and $\mathcal{P}=\{0.4,0.2,0.1,0.1,0.1,0.1\}$, and use $r=t-n$ as the reference. 
Results in~\cref{tab:gap_ablation} confirm consistent gains in heatmap quality and localization stability over uniform and random sampling.

\section{Conclusion}

We introduce \textbf{SurgAtt-1.16M}, a large-scale benchmark unified from multiple sources for studying surgical attention tracking across diverse organs and conditions, and propose \textbf{SurgAtt-Tracker}, an holistic framework that tracks attention via proposal reranking and motion-aware refinement, achieving state-of-the-art results. Limitations include reliance on detector proposal coverage, potential drift under prolonged occlusions or abrupt camera motion, and a current focus on attention estimation rather than camera control. Future work will integrate SurgAtt-Tracker into robotic systems for safe, stable, and intent-aware endoscope control.

\section*{Impact Statement}

This research bridges a critical gap in medical robotics by providing the first large-scale benchmark to enable attention-aware endoscopic control. By open-sourcing SurgAtt-1.16M, we aim to democratize access to high-quality clinical data, accelerating the development of intelligent surgical assistants that can reduce procedural variability and cognitive fatigue. While our framework advances surgical autonomy, we emphasize that attention tracking is probabilistic; clinical deployment requires rigorous ``human-in-the-loop'' validation to prevent potential misguidance during rare adverse events. All patient data in this study was strictly de-identified to safeguard privacy.

\nocite{langley00}

\bibliography{example_paper}
\bibliographystyle{icml2026}

\newpage
\appendix
\onecolumn
\section{Appendix}

\subsection{Dataset Processing and Detailed Supplementation\label{app:dataset_appendix}}

\begin{figure*}[h]
  \begin{center}
\centerline{\includegraphics[width=\textwidth]{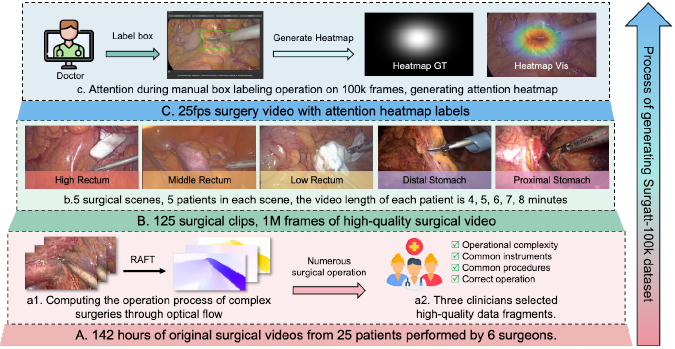}}
    \caption{
      \textbf{Construction pipeline of the SurgAtt-SZPH dataset.}
Raw laparoscopic videos are curated into high-quality surgical clips via optical-flow–based operation analysis and expert screening. Videos are sampled at 25 fps and grouped into five representative surgical scenes. During annotation, surgeons mark attention regions with bounding boxes, which are converted into continuous attention heatmaps for supervision. The resulting dataset provides dense, high-fidelity attention annotations across diverse surgical scenarios.
    }
    \label{fig:appendix_dataset}
    \vskip -0.3in
  \end{center}
\end{figure*}

\subsubsection{Dataset Annotation Pipline}
The curation process is illustrated in~\cref{fig:appendix_dataset} and consists of two stages to efficiently obtain informative and clinically meaningful attention supervision.

\paragraph{Stage 1 (a1): motion-based mining} Starting from 142 hours of raw endoscopic videos, we run RAFT optical flow to estimate frame-to-frame motion and discard temporally static or low-motion segments that typically contain limited viewpoint changes and weak attention transitions. This step narrows the search space to segments with moderate activity, improving annotation efficiency while preserving diverse camera motions and instrument–tissue interactions.

\paragraph{Stage 2 (a2): clinician screening} Candidate segments are reviewed by expert surgeons to ensure clinical relevance and data quality, using four criteria: (1) operational complexity, (2) coverage of common surgical instruments, (3) inclusion of representative procedures, and (4) correctness of surgical operations. We then sample 30 minutes per patient and split them into clips of 4–8 minutes to balance temporal continuity with manageable clip length. The resulting SurgAtt-SZPH subset covers five surgical sites (High/Middle/Low Rectum, Distal/Proximal Stomach) and contains 125 clips with about 1M frames at 25 fps.

\paragraph{Annotation motivation} Rather than exhaustively segmenting objects, surgeons label a single box that captures the clinically attended region (e.g., the active dissection area or target anatomy) per frame, which is then converted into a normalized attention heatmap. This design (i) aligns supervision with surgeon intent and viewpoint control, (ii) yields dense, temporally coherent targets suitable for tracking and heatmap evaluation, and (iii) substantially reduces annotation cost compared with pixel-level labeling while remaining robust to appearance ambiguity in endoscopic scenes.

\begin{table*}[h]
  \caption{\textbf{SurgAtt-1.16M} forms a complete benchmark by aggregating three subsets under a unified annotation protocol, and summarizes the data source, organ, procedure, image resolution, and the number of frames in the train/test splits.}
  \label{tab:surgatt_datasets}
  \centering
    \begin{center}
      \begin{sc}
  \resizebox{\textwidth}{!}{
  \begin{tabular}{l l l l r r r r}
    \toprule
    \textbf{Dataset Name} & \textbf{Data Source} & \textbf{Surgical Organ} & \textbf{Procedure} & \textbf{Image Size} & \textbf{Train} & \textbf{Test} & \textbf{Total} \\
    \midrule
    SurgAtt-SZPH            & Anonymous Hospital  & Rectum, Stomach & Rectal and gastric cancer resections & 1920x1080    & 800,000 & 200,000 & 1,000,000 \\
    SurgAtt-AutoLaparo              & AutoLaparo             & Uterus          &  Hysterectomy                  & 1920x1080       & 60,000  & 15,000  & 75,000    \\
    SurgAtt-Hamlyn          & Hamlyn Centre           & Kidney          & Nephrectomy                    & 384x192        & 68,478  & 14,380  & 82,858    \\
        \midrule

    \textbf{SurgAtt-1.16M} & Aggregated         & Multiple        & Multiple laparoscopic procedures     & Multiple               & 928,478 & 229,380 & 1,157,858 \\
    \bottomrule
  \end{tabular}
  }
        \end{sc}
  \end{center}
\end{table*}

\begin{table}[h]
  \caption{Two-stage curation and annotation pipeline for the SurgAtt-SZPH (SurgAtt-1M) subset. We first mine informative surgical segments via optical-flow–based motion filtering (A), then clinicians select high-quality clips under four clinical criteria (B), and finally surgeons annotate attention boxes that are converted into attention heatmaps at 25 fps across five surgical sites (C).}
  \label{tab:dataset_stats}
  \begin{center}
    \begin{small}
      \begin{sc}
        \resizebox{0.9\textwidth}{!}{
        \begin{tabular}{lccccccccccc}
          \toprule
          Category 
          & \multicolumn{2}{c}{Distal Gastric}
          & \multicolumn{2}{c}{Proximal Gastric}
          & \multicolumn{2}{c}{High Rectum}
          & \multicolumn{2}{c}{Middle Rectum}
          & \multicolumn{2}{c}{Low Rectum}
          & Total \\
          \cmidrule(lr){1-11}
          Split & Train & Test & Train & Test & Train & Test & Train & Test & Train & Test &  \\
          \midrule
          Patients & 4 & 1 & 4 & 1 & 4 & 1 & 4 & 1 & 4 & 1 & 25 \\
          Videos & 20 & 5 & 20 & 5 & 20 & 5 & 20 & 5 & 20 & 5 & 125 \\
          Frames & 180K & 45K & 180K & 45K & 180K & 45K & 180K & 45K & 180K & 45K & 1.12M \\
          \bottomrule
        \end{tabular}
        }
      \end{sc}
    \end{small}
  \end{center}
\end{table}

\subsubsection{SurgAtt-1.16M Dataset Details Supplement.}
\paragraph{Dataset statistics and composition.}
\cref{tab:dataset_stats} reports detailed statistics of the \textbf{SurgAtt-SZPH} subset, stratified by five surgical sites (Distal/Proximal Gastric and High/Middle/Low Rectum). We adopt a \textbf{patient-level split} for each site to avoid identity leakage: each site contains 5 patients, with 4 patients for training and 1 patient for testing. In total, SurgAtt-SZPH includes 25 patients and 125 surgical clips. The clips are obtained by sampling comparable-duration segments per patient and segmenting them into multiple shorter videos, yielding a balanced number of clips per site (Train/Test = 20/5) and a near-uniform frame budget per site (Train/Test = 180K/45K), for an overall size of \textbf{1.12M} frames. These balanced per-site statistics enable controlled evaluation of attention tracking under different anatomical regions and surgical contexts.

\paragraph{Full benchmark aggregation.}
\cref{tab:surgatt_datasets} summarizes the full \textbf{SurgAtt-1.16M} benchmark, which \textbf{aggregates} SurgAtt-SZPH with re-labeled versions of \textbf{AutoLaparo} and \textbf{Hamlyn}. Specifically, AutoLaparo contributes uterus/hysterectomy cases, and Hamlyn provides kidney/nephrectomy cases with different visual characteristics and resolutions. We re-annotate these external sources under the \textbf{same attention definition and heatmap generation protocol} as SurgAtt-SZPH, so that all subsets can be evaluated with a unified metric suite and a consistent preprocessing pipeline. As a result, SurgAtt-1.16M expands organ/procedure coverage while preserving annotation consistency, supporting both in-domain studies on SurgAtt-SZPH and cross-domain generalization evaluation on heterogeneous surgical settings.

\subsection{Heatmap Generation Details}
\label{app:heatmap_generation}

\begin{figure*}[t]
  \begin{center}
\centerline{\includegraphics[width=0.8\textwidth]{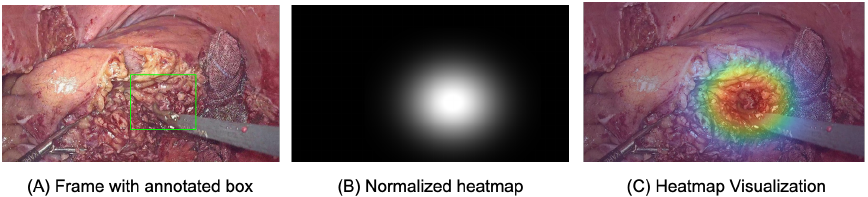}}
    \caption{
     Visualization of heatmap generation.
    }
    \label{fig:appendix_heatmap}
  \end{center}
  \vskip -0.3in
\end{figure*}

Given a discrete sequence of per-frame attention boxes, we convert box annotations into dense ground-truth attention maps by (i) mapping each box to a box-aligned anisotropic Gaussian kernel, (ii) temporally accumulating kernels with exponential decay to simulate visual persistence, and (iii) applying smoothing and robust normalization to obtain a bounded heatmap in $[0,1]$, as shown in~\cref{fig:appendix_heatmap}.

\paragraph{Box-to-kernel mapping.}
For a frame $t$, each annotated box $B_t^i=(c_x,c_y,w,h)$ (center, width, height; in pixels after resizing to the evaluation resolution) is mapped to an anisotropic Gaussian kernel
$G(\mathbf{x};\boldsymbol{\mu}_t^i,\boldsymbol{\Sigma}_t^i)$,
with mean $\boldsymbol{\mu}_t^i=(c_x,c_y)$ and diagonal covariance
$\boldsymbol{\Sigma}_t^i=\mathrm{diag}(\sigma_x^2,\sigma_y^2)$.
Following our implementation, we set
$\sigma_x=\max(1,\,s\,w)$ and $\sigma_y=\max(1,\,s\,h)$ with scale factor $s$ (default $s=0.45$),
and evaluate the kernel within a truncated window of $\pm 3\sigma$ along each axis for efficiency.
When multiple boxes are present in the same frame, we aggregate them by summation:
\begin{equation}
\tilde{G}_t(\mathbf{x}) \;=\; \sum_i \, a(w_i,h_i)\; G(\mathbf{x};\boldsymbol{\mu}_t^i,\boldsymbol{\Sigma}_t^i),
\label{eq:gt_sum}
\end{equation}
where $a(w,h)$ is an optional \emph{area compensation} term to reduce scale bias.
By default, we use $a(w,h)=1/\sqrt{\max(wh,1)}$ (``sqrt''), which prevents large boxes from dominating the heatmap solely due to their area.

\paragraph{Temporal accumulation (visual persistence).}
We simulate human visual persistence by exponentially decaying the previous density map and adding the current-frame kernel:
\begin{equation}
\mathbf{M}_t(\mathbf{x})
\;=\;
(1-\alpha)\,\mathbf{M}_{t-1}(\mathbf{x}) \;+\; \tilde{G}_t(\mathbf{x}),
\qquad
\mathbf{M}_0=\mathbf{0},
\label{eq:mt_decay}
\end{equation}
where $\alpha\in(0,1)$ controls memory length (default $\alpha=0.22$).
If frame $t$ has no valid annotation, we simply apply the decay term, i.e., $\tilde{G}_t\equiv 0$.

\paragraph{Post-processing and robust normalization.}
To reduce high-frequency artifacts and improve spatial coherence, we apply Gaussian smoothing with kernel size $k$ (default $k=9$):
\begin{equation}
\mathbf{S}_t \;=\; \mathcal{G}(\mathbf{M}_t; k).
\label{eq:smooth}
\end{equation}
We then normalize by a robust upper bound using the $p$-th percentile
$Q_p(\mathbf{S}_t)$ (default $p=99.5$) to suppress outliers and stabilize dynamic range:
\begin{equation}
\mathbf{H}_t
\;=\;
\mathrm{clip}\!\left(
\frac{\mathbf{S}_t}{\max(Q_p(\mathbf{S}_t),\epsilon)},
\,0,\,1
\right),
\label{eq:robust_norm}
\end{equation}
where $\epsilon$ is a small constant.
The resulting $\mathbf{H}_t\in[0,1]$ is stored as a grayscale map and optionally visualized by overlaying a colorized heatmap on the resized frame.

\paragraph{Implementation notes.}
All boxes are parsed from YOLO-format labels (normalized $(c_x,c_y,w,h)$) and converted to pixel coordinates under a fixed output resolution (default $960\times540$). The kernel is peak-normalized and truncated to $\pm3\sigma$, and we optionally inflate the box scale prior to kernel generation (default inflation $0$). Unless stated otherwise, we use $\alpha=0.22$, $s=0.45$, $k=9$, and $p=99.5$ throughout.

\subsection{Evaluation Metric Definitions\label{app:experiment_detail}}

Let $P\in[0,1]^{H\times W}$ be the predicted attention heatmap and $G\in[0,1]^{H\times W}$ the ground-truth heatmap for a frame, both normalized to $[0,1]$. 
Let $N=HW$ and denote by $P_i$ and $G_i$ the $i$-th pixel after vectorization.

\paragraph{MAE.}
\begin{equation}
\mathrm{MAE}(P,G)=\frac{1}{N}\sum_{i=1}^{N}\left|P_i-G_i\right|.
\end{equation}

\paragraph{MSE.}
\begin{equation}
\mathrm{MSE}(P,G)=\frac{1}{N}\sum_{i=1}^{N}\left(P_i-G_i\right)^2.
\end{equation}

\paragraph{CC (Pearson correlation coefficient).}
Let $\mu_P=\frac{1}{N}\sum_{i=1}^{N}P_i$ and $\mu_G=\frac{1}{N}\sum_{i=1}^{N}G_i$. Then
\begin{equation}
\mathrm{CC}(P,G)=
\frac{\sum_{i=1}^{N}(P_i-\mu_P)(G_i-\mu_G)}
{\sqrt{\sum_{i=1}^{N}(P_i-\mu_P)^2}\ \sqrt{\sum_{i=1}^{N}(G_i-\mu_G)^2}+\epsilon},
\end{equation}
where $\epsilon$ is a small constant for numerical stability.

\paragraph{SIM (histogram intersection).}
We first form $\ell_1$-normalized non-negative maps:
\begin{equation}
\hat{P}_i=\frac{\max(P_i,0)}{\sum_{j=1}^{N}\max(P_j,0)+\epsilon},\qquad
\hat{G}_i=\frac{\max(G_i,0)}{\sum_{j=1}^{N}\max(G_j,0)+\epsilon}.
\end{equation}
If the denominator is zero (i.e., the map sums to zero), we use a uniform distribution $\hat{P}_i=\frac{1}{N}$ (and similarly for $\hat{G}$).
SIM is then defined as
\begin{equation}
\mathrm{SIM}(P,G)=\sum_{i=1}^{N}\min(\hat{P}_i,\hat{G}_i).
\end{equation}

\paragraph{NSS (Normalized Scanpath Saliency).}
For NSS, we follow a standard saliency protocol by z-normalizing the prediction and averaging scores on ground-truth salient locations, where salient locations are defined as the top $5\%$ pixels of the ground-truth heatmap. We z-normalize the prediction map:
\begin{equation}
P_i^{z}=\frac{P_i-\mu_P}{\sigma_P+\epsilon},\qquad
\sigma_P=\sqrt{\frac{1}{N}\sum_{i=1}^{N}(P_i-\mu_P)^2}.
\end{equation}
Since our supervision provides dense heatmaps, we construct a binary salient-location mask $F\in\{0,1\}^{H\times W}$ from $G$ by thresholding at the per-image top-quantile:
\begin{equation}
F_i=\mathbb{I}\!\left[G_i\geq Q_{q}(G)\right],
\end{equation}
where $Q_{q}(G)$ denotes the $q$-quantile of pixel values in $G$ and we use $q=0.95$ (top $5\%$ pixels) in all experiments.
Finally,
\begin{equation}
\mathrm{NSS}(P,G)=\frac{1}{\sum_{i=1}^{N}F_i}\sum_{i=1}^{N}P_i^{z}\,F_i,
\end{equation}
with the denominator clamped to at least 1 for numerical stability.

\begin{algorithm}[h]
  \caption{Unified Training and Inference of SurgAtt-Tracker}
  \label{alg:unified_train_infer}
  \begin{algorithmic}[1]
    \STATE \textbf{Input:} frames $\{I_t\}_{t=1}^{T}$; \textbf{training only:} GT boxes $\{B_t^{\mathrm{gt}}\}$
    \STATE \textbf{Mode:} $\mathsf{mode}\in\{\textsc{train},\textsc{infer}\}$
    \STATE \textbf{Train gaps:} $\mathcal{N}=\{1,2,4,8,16,32\}$, \ \ $\mathcal{P}=\{0.4,0.2,0.1,0.1,0.1,0.1\}$
    \STATE Initialize $\hat{B}_1$ (e.g., detector Top-1 on $I_1$)

    \FOR{$t=2$ \textbf{to} $T$}
        \IF{$\mathsf{mode}=\textsc{train}$}
            \STATE $n \leftarrow \mathrm{SampleGap}(\mathcal{N},\mathcal{P})$, \ \ $r \leftarrow t-n$
        \ELSE
            \STATE $n \leftarrow 1$, \ \ $r \leftarrow t-1$
        \ENDIF

        \STATE $(\mathcal{B}_{r}, \mathcal{F}_{r}^{y}) \leftarrow D(I_{r})$, \ \ $(\mathcal{B}_{t}, \mathcal{F}_{t}^{y}) \leftarrow D(I_{t})$

        \IF{$\mathsf{mode}=\textsc{train}$}
            \STATE $B_r^{s} \leftarrow \mathrm{BoxSelection}(\mathcal{B}_{r}, B_{r}^{\mathrm{gt}})$
        \ELSE
            \STATE $B_r^{s} \leftarrow \hat{B}_{t-1}$
        \ENDIF

        \STATE $f_r \leftarrow \mathrm{MSR}(\mathcal{F}_{r}^{y}, B_r^{s})$
        \STATE $\{f_t^k\}_{k=1}^{K} \leftarrow \mathrm{MSR}(\mathcal{F}_{t}^{y}, \mathcal{B}_{t})$

        \STATE $\{s_t^k\}_{k=1}^{K} \leftarrow \mathrm{AS\mbox{-}Rerank}(f_r, \{f_t^k\}_{k=1}^{K})$
        \STATE $\hat{k}_t \leftarrow \arg\max_k s_t^k$
        \STATE $\hat{B}_t \leftarrow B_t^{\hat{k}_t}$, \ \ $\hat{f}_t \leftarrow f_t^{\hat{k}_t}$

        \STATE $B_t^{r} \leftarrow \mathrm{MAA\mbox{-}Refine}(\hat{f}_t, \mathrm{Geo}(\hat{B}_t,B_r^{s}))$
        \STATE $H_t \leftarrow \mathrm{Heat}(\mathcal{F}_t^{y}, B_t^{r})$
    \ENDFOR
  \end{algorithmic}
\end{algorithm}

\subsection{Unified Training and Inference Pipeline}
\label{app:unified_train_infer}

Alg.~\ref{alg:unified_train_infer} summarizes the end-to-end pipeline used by \textbf{SurgAtt-Tracker} for both training and online inference.
Given a video clip $\{I_t\}_{t=1}^{T}$, a frozen detector $D(\cdot)$ is applied to each queried frame to produce (i) a Top-$K$ proposal set $\mathcal{B}_t=\{B_t^k\}_{k=1}^{K}$ and (ii) multi-scale neck features $\mathcal{F}_t^{y}$.
The tracker then uses a reference frame $r$ and a reference box $B_r^{s}$ to compute a reference ROI embedding $f_r$, which conditions proposal reranking on the target frame $t$.

\paragraph{Training gap sampling.}
Using only adjacent pairs ($r=t-1$) during training may over-emphasize near-static transitions and encourage a shortcut where the refiner behaves almost as identity, i.e., simply propagating the previous spatial prior while still achieving low loss.
To expose the model to diverse displacements and fast tool motion, we sample the temporal gap $n$ from a categorical distribution $n\sim \mathrm{Cat}(\mathcal{N},\mathcal{P})$ with $\mathcal{N}=\{1,2,4,8,16,32\}$ and $\mathcal{P}=\{0.4,0.2,0.1,0.1,0.1,0.1\}$, and set $r=t-n$ in training.

\paragraph{Training mode.}
For each target frame $t\ge2$, we sample a gap $n$ and set $r=t-n$ (clipped to the valid range if necessary).
We run the detector on both frames $I_r$ and $I_t$ to obtain $(\mathcal{B}_r,\mathcal{F}_r^{y})$ and $(\mathcal{B}_t,\mathcal{F}_t^{y})$.
To avoid teacher-forcing mismatch while keeping the optimization stable, we construct the reference box $B_r^{s}$ via an oracle selection on frame $r$:
\emph{among valid detector proposals, choose the one with minimum center error to the ground-truth box $B_r^{\mathrm{gt}}$}.
Then we extract the reference ROI embedding $f_r=\mathrm{MSR}(\mathcal{F}_r^{y}, B_r^{s})$ and the target proposal embeddings $\{f_t^k\}_{k=1}^{K}=\mathrm{MSR}(\mathcal{F}_t^{y},\mathcal{B}_t)$ using the same multi-scale ROI extractor $\mathrm{MSR}(\cdot)$.
The reranker $\mathrm{AS\mbox{-}Rerank}$ outputs scores $\{s_t^k\}_{k=1}^{K}$ conditioned on $f_r$, from which we select $\hat{k}_t=\arg\max_k s_t^k$ and obtain the Top-1 box $\hat{B}_t$ and its embedding $\hat{f}_t$.
Finally, the refiner $\mathrm{MAA\mbox{-}Refine}$ predicts the refined box $B_t^{r}$ using the selected embedding and a motion-geometry descriptor $\mathrm{Geo}(\hat{B}_t,B_r^{s})$.
All losses are computed on frame $t$ only (reranking supervision and refinement regression), while $r$ serves as a reference for conditioning.

\paragraph{Inference mode.}
At test time, the same modules are executed online with a one-step reference ($n=1$, $r=t-1$), and the reference box is set to the previous prediction, $B_r^{s}\leftarrow \hat{B}_{t-1}$.
This matches the realistic tracking condition where only past predictions are available.
We then rerank the Top-$K$ proposals on $I_t$, select $\hat{B}_t$, refine it to $B_t^{r}$, and optionally render the attention heatmap $H_t=\mathrm{Heat}(\mathcal{F}_t^{y},B_t^{r})$.

\subsection{Additional Experimental Setting and Results}

\subsubsection{Experimental Setting}
\paragraph{Baseline implementation details.}
To ensure a fair and consistent comparison, we follow a unified evaluation protocol and adopt minimal, task-aligned adaptations for different categories of baselines.

\textbf{U-Net--based methods.}
For U-Net--based attention models, we strictly preserve the original network architectures and do not introduce any structural modifications. The only change is to replace the original loss with a binary cross-entropy (BCE) loss computed against the ground-truth attention heatmaps. All models are trained to directly predict a single-channel, normalized heatmap in $[0,1]$, which is evaluated under the same metric suite as our method.

\textbf{Regression-based methods.}
For regression-based baselines, we do not alter either the network architecture or the original loss formulation. These models are trained and evaluated directly on our dataset following their default settings, with their predicted outputs converted into normalized attention heatmaps when necessary for metric computation.

\textbf{Object-tracking--based methods.}
For object-tracking approaches, we adopt a YOLO-style annotation format to ensure training consistency across datasets. During inference, we provide the ground-truth bounding box of the first frame as initialization, following standard tracking evaluation protocols. The tracker then propagates bounding boxes over time, and the resulting box sequences are converted into attention heatmaps using the same box-to-heatmap generation procedure as described in \cref{app:heatmap_generation}. These heatmaps are evaluated using identical attention metrics.

\textbf{Detection-based methods.}
Detection-based baselines are trained using the YOLO format without architectural modification. At inference time, detected bounding boxes are transformed into attention heatmaps using the same generation strategy as above, enabling direct comparison with tracking- and heatmap-based approaches under a unified evaluation framework.

\begin{figure*}[t]
  \begin{center}
\centerline{\includegraphics[width=0.9\textwidth]{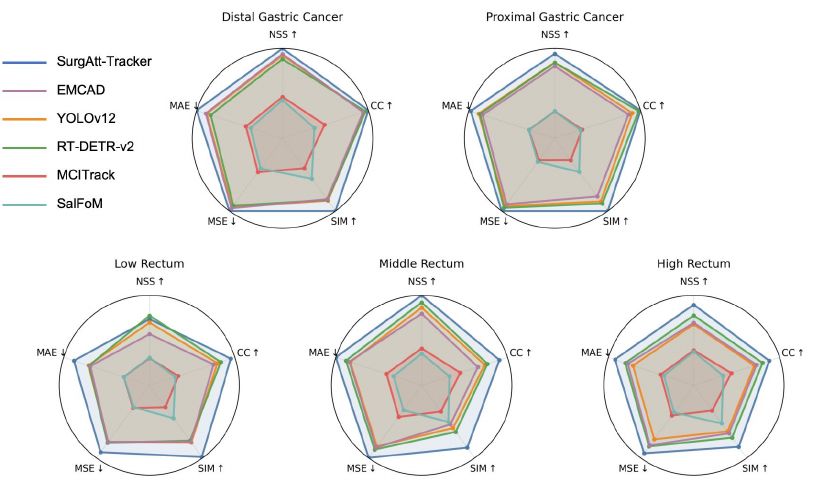}}
    \caption{
      Radar plots on five subsets (Distal/Proximal Gastric Cancer, Low/Middle/High Rectum) comparing representative SOTA methods over NSS/CC/SIM (higher better) and MSE/MAE (lower better).
    }
    \label{fig:5_split_szph}
  \end{center}
  \vskip -0.3in
\end{figure*}

\begin{figure*}[t]
  \begin{center}
\centerline{\includegraphics[width=\textwidth]{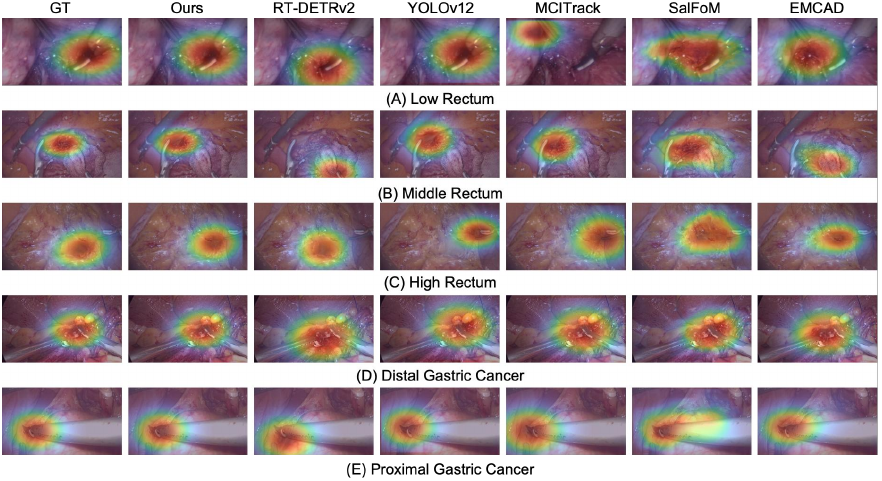}}
    \caption{Qualitative comparison of attention heatmap predictions on SurgAtt-SZPH across five surgical scenarios. SurgAtt-Tracker produces sharper and more stable attention aligned with clinically relevant regions compared with representative SOTA baselines.}
    \label{fig:app_szph}
  \end{center}
  \vskip -0.3in
\end{figure*}

\begin{figure*}[t]
  \begin{center}
\centerline{\includegraphics[width=\textwidth]{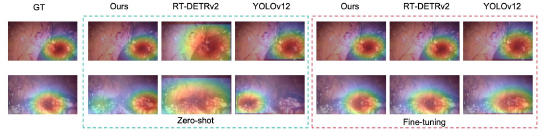}}
   \caption{Qualitative comparison on SurgAtt-Hamlyn under zero-shot and fine-tuning settings. SurgAtt-Tracker shows more stable and better-localized attention than RT-DETRv2 and YOLOv12 across both regimes.}
    \label{fig:vis_hamlyn}
  \end{center}
  \vskip -0.3in
\end{figure*}

\begin{figure*}[t]
  \begin{center}
\centerline{\includegraphics[width=\textwidth]{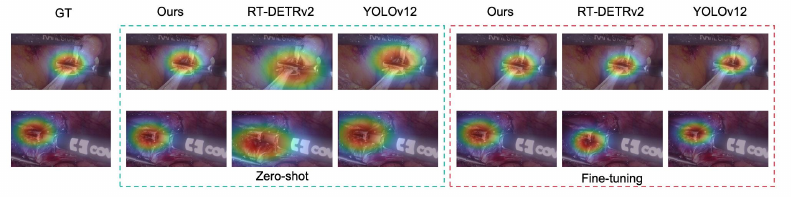}}
    \caption{Qualitative comparison on SurgAtt-AutoLaparo under zero-shot and fine-tuning settings. Our method maintains compact and consistent attention maps under domain shift and further improves after fine-tuning.}
    \label{fig:vis_autolaparo}
  \end{center}
  \vskip -0.3in
\end{figure*}

\subsubsection{Main Experiment Visualization and Analysis\label{app:szph_scene_radar}}

\paragraph{Per-scene analysis on SurgAtt-SZPH.}
\cref{fig:5_split_szph} visualizes a per-scene comparison across five SurgAtt-SZPH subsets, covering two gastric cancer scenes (distal/proximal) and three rectum scenes (low/middle/high).
Each radar plot summarizes five complementary metrics: NSS, CC, and SIM evaluate alignment between predicted and ground-truth attention distributions (higher is better), while MSE and MAE measure pixel-wise heatmap errors (lower is better).
Across all scenes, \textbf{SurgAtt-Tracker} consistently forms the outer envelope on NSS/CC/SIM while remaining closest to the optimal low-error region on MSE/MAE, indicating both stronger correlation and sharper localization.
Notably, the improvements persist under diverse scene characteristics (e.g., varying anatomy, viewpoint changes, and instrument interactions), suggesting that our temporal-consistent reranking and motion-aware refinement generalize reliably beyond any single surgical condition.

\paragraph{Qualitative analysis on SurgAtt-SZPH.}
\cref{fig:vis_res} presents qualitative comparisons of attention heatmap predictions on the SurgAtt-SZPH dataset across five representative surgical scenarios: Low Rectum, Middle Rectum, High Rectum, Distal Gastric Cancer, and Proximal Gastric Cancer. Each row shows the ground-truth attention heatmap and predictions from representative detection-, tracking-, saliency-, and segmentation-based baselines, together with our method.

Across all scenarios, \textbf{SurgAtt-Tracker} produces attention maps that are both spatially concentrated and well aligned with clinically relevant regions. In rectal surgery scenes (A--C), where attention often shifts subtly along tissue boundaries or instrument--tissue interaction zones, our predictions closely match the ground truth in both location and extent. In contrast, detection-based methods (e.g., YOLOv12, RT-DETRv2) tend to generate coarse or overly centralized responses that fail to adapt to fine-grained attention shifts, while object-tracking methods (e.g., MCITrack) may drift toward instruments or background regions when visual cues become ambiguous.

In gastric cancer scenarios (D--E), which typically involve larger anatomical structures and more pronounced camera motion, saliency- and segmentation-based methods (e.g., SalFoM, EMCAD) often produce diffuse or over-smoothed heatmaps, spreading attention across broad regions and reducing localization precision. Our method maintains compact, single-mode attention distributions that remain centered on the surgical target despite specular highlights, smoke, or partial occlusions. This robustness is attributed to the combination of proposal-level temporal reranking, which stabilizes attention selection over time, and motion-aware refinement, which constrains updates under rapid viewpoint changes.

Overall, these qualitative results are consistent with the quantitative improvements reported in the main paper and appendix. They demonstrate that modeling surgical attention as temporally consistent retrieval over a high-recall proposal set, followed by motion-conditioned continuous refinement, yields attention heatmaps that are sharper, more stable, and more clinically interpretable across diverse surgical scenes.

\paragraph{Zero-shot and fine-tuning generalization analysis.\label{app:other_data}}
\cref{fig:vis_hamlyn,fig:vis_autolaparo} present qualitative comparisons on the \textbf{SurgAtt-Hamlyn} and \textbf{SurgAtt-AutoLaparo} datasets, evaluating three representative SOTA methods (RT-DETRv2, YOLOv12, and our approach) under both zero-shot and fine-tuning settings. These datasets differ substantially from SurgAtt-SZPH in organ type, surgical procedure, imaging resolution, and visual appearance, providing a challenging testbed for cross-domain generalization.

In the \textbf{zero-shot} setting, SurgAtt-Tracker demonstrates strong robustness to domain shifts. On SurgAtt-Hamlyn, which involves kidney surgery with smaller target regions and frequent instrument occlusions, our method consistently produces compact and well-localized attention heatmaps that remain aligned with the clinically relevant regions. In contrast, detection-based baselines often exhibit either spatial drift or over-smoothed responses, particularly when specular highlights or motion blur dominate the scene. Similar trends are observed on SurgAtt-AutoLaparo, where changes in organ texture and instrument appearance cause competing methods to spread attention across large regions, while our predictions remain stable.

Under the \textbf{fine-tuning} setting, all methods benefit from domain-specific supervision; however, notable qualitative differences persist. While RT-DETRv2 and YOLOv12 improve localization accuracy, their predictions often remain sensitive to background clutter or instrument dominance. SurgAtt-Tracker further sharpens attention boundaries and maintains temporal consistency, producing single-mode heatmaps that better match the ground truth across diverse frames. This suggests that the gains of our approach stem not only from fine-tuning but also from its design principles—namely, proposal-level temporal reranking and motion-aware refinement, which explicitly model attention evolution over time.

These qualitative results indicate that SurgAtt-Tracker generalizes more effectively across datasets in the zero-shot regime and continues to provide stable, well-aligned attention estimates after fine-tuning. The observed behavior is consistent with our quantitative results and supports the effectiveness of decoupling attention selection from motion-conditioned refinement when transferring across heterogeneous surgical domains.

\subsection{Additional Ablation Study \label{app:ablation}}

\begin{figure}[h]
  \begin{center}
\centerline{\includegraphics[width=0.5\linewidth]{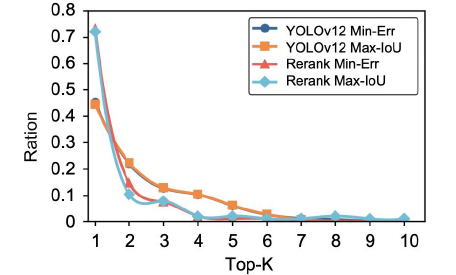}}
    \caption{Top-$K$ analysis of proposal ranking quality. AS-Rerank reorders the same YOLOv12 candidate pool using temporal consistency, increasing the likelihood that oracle-best proposals (Min-Err / Max-IoU) appear in the first few ranks.}
    \label{fig:topk_ratio}
  \end{center}
  \vskip -0.3in
\end{figure}

\paragraph{AS-Rerank contributes.}
We adopt a strong detector (YOLOv12) as a \emph{high-recall proposal generator} rather than a final decision maker. In surgical scenes, detector confidence is often poorly calibrated to clinical attention quality due to specular highlights, smoke, blood occlusion, and frequent appearance changes of instruments and tissues. As a result, selecting the Top-1 box purely by confidence can miss the most attention-aligned region, even when it is already present in the Top-$K$ candidate set.

\cref{fig:topk_ratio} provides evidence for this design choice. The plot reports, for each Top-$K$, the fraction of frames where the \emph{best} proposal (defined by oracle criteria such as minimum center error or maximum IoU w.r.t.\ the annotated box) appears within the top ranks. We observe that YOLOv12 indeed produces high-quality hypotheses early in the list, but these hypotheses are not consistently ranked as Top-1 under confidence. After applying AS-Rerank, the same candidate pool is reordered by temporal consistency, which increases the likelihood that the oracle-best proposal is promoted into the first few ranks (especially Top-1/Top-2). This directly explains why the reranking-only variant improves both localization quality and downstream heatmap metrics without changing the detector backbone: AS-Rerank does not create new boxes; it \emph{recovers better hypotheses already present in the candidate set} by leveraging temporal coherence. 

Finally, building on the reranked Top-1 hypothesis, MAA-Refine further reduces residual localization error by conditioning the correction on motion cues from the reference state, yielding additional gains beyond reranking alone. Together, these results justify our two-stage design: a high-recall detector to provide diverse candidates, temporal reranking to select the most consistent attention hypothesis, and motion-aware refinement to obtain a precise and stable final prediction.



\end{document}